\documentclass{ifacconf}

\usepackage{graphicx}      
\usepackage{amsmath}
\usepackage{amsfonts}
\usepackage{natbib}        

\usepackage{todonotes}
\usepackage{blkarray}
\usepackage{tabularx}
\usepackage{colortbl}
\usepackage{hhline}
\usepackage{multirow}

\newcommand{\R}{\mathbb{R}}

\newcommand{\Z}{\mathbb{Z}}

\begin{document}
\begin{frontmatter}

\title{Chatter Classification in Turning using Machine Learning and Topological Data Analysis\thanksref{footnoteinfo}}

\thanks[footnoteinfo]{This material is based upon work supported by the National Science Foundation under Grant Nos.~CMMI-1759823 and DMS-1759824 with PI FAK,
and CMMI-1800466 and DMS-1800446 with PI EM. JAP acknowledges the support of the NSF  under grant DMS-1622301 and  DARPA under grant HR0011-16-2-003.
}

\author[Firas]{Firas A.~Khasawneh}
\author[Liz]{Elizabeth Munch}
\author[Jose]{Jose A. Perea}

\address[Firas]{Dept.~of Mechanical Engineering (e-mail: khasawn3@egr.msu.edu)}

\address[Liz]{Dept.~of Computational Mathematics, Science, and Engineering; Dept.~of Mathematics(e-mail: muncheli@egr.msu.edu)}
\address[Jose]{Dept.~of Computational Mathematics, Science, and Engineering; Dept.~of Mathematics (e-mail: joperea@msu.edu)}
\address{Michigan State University, East Lansing, MI 48824 USA }

\begin{abstract}                
Chatter identification and detection in machining processes has been an active area of research in the past two decades.
Part of the challenge in studying chatter is that machining equations that describe its occurrence are often nonlinear delay differential equations.
The majority of the available tools for chatter identification rely on defining a metric that captures the characteristics of chatter, and a threshold that signals its occurrence.
The difficulty in choosing these parameters can be somewhat alleviated by utilizing machine learning techniques.
However, even with a successful classification algorithm, the transferability of typical machine learning methods from one data set to another remains very limited.
In this paper we combine supervised machine learning with Topological Data Analysis (TDA) to obtain a descriptor of the process which can detect chatter.
The features we use are derived from the persistence diagram of an attractor reconstructed from the time series via  Takens embedding.
We test the approach using deterministic and stochastic turning models, where the stochasticity  is introduced via the cutting coefficient term.
Our results show a $97\%$ successful classification rate on the deterministic model labeled by the stability diagram obtained using the spectral element method.
The features gleaned from the deterministic model are then utilized for characterization of chatter in a stochastic turning model where there are very limited analysis methods.
\end{abstract}


\end{frontmatter}


\section{Introduction}
Cutting processes such as turning and milling, which are often described using delay differential equations, constitute a major part of discrete manufacturing.
One of the most intriguing problems that face the industry in these processes is chatter.
Chatter is characterized by large amplitude oscillations that have detrimental effects on the workpiece quality, cutting tool, and the spindle.
Therefore, a large body of research has emerged focused on predicting and modeling chatter.

In addition to predictive models for chatter, prior works have  studied in-situ methods for chatter detection (\cite{Smith1992,Altintas1992,Choi2003,Kuljanic2009,Yao2010,Elias2014}).
These methods are often combined with active control strategies to suppress chatter (\cite{Faassen2007, Dijk2010}).
In-process chatter detection methods attempt to alleviate the process modeling difficulties by investigating the cutting signal itself.
Currently available in-process methods for studying chatter typically rely on comparing the characteristics of the acoustic, vibration, or force signals against certain predefined features indicative of chatter (\cite{Tlusty1983,Delio1992,Gradisek1998a,Schmitz2002,Choi2003,Bediaga2009,Sims2009,Nair2010,Dijk2010,Tsai2010,Kakinuma2011,Ma2013}).
Often a metric or an index and a threshold are defined on the time series and chatter is detected if this threshold is exceeded.
However, online chatter detection is associated with another set of challenges.
For example, the value of an adequate threshold is difficult to identify and the majority of tools used to define the necessary metrics on the data are linear despite the inherently nonlinear nature of machining processes.
Further, each specific case of cutting conditions would require performing preliminary cutting tests to train the used algorithms (\cite{Hino2006}).
Another difficulty is the need for accurately identifying threshold values to compare certain features in the signal against those associated with chatter (\cite{Choi2003}).
Even if a suitable threshold is found, many online chatter detection methods indicate the occurrence of chatter only after it has fully developed, hence damaging the workpiece or the tool (\cite{Faassen2007}).
Chatter-detection using currently available classification and machine learning methods requires a training set for each setup of the process.
Consequently, if data is obtained for the same type of cutting process but with a different configuration, a new training set is generally required.

In contrast, this paper seeks to apply learning on features of a topological descriptor of the signal, rather than to features of the signal itself, thus basing the learning on intrinsic process characteristics which can be generalized and adapted to different situations.
The tools we use come from Topological Data Analysis (TDA) and nonlinear time series analysis.
Specifically, we embed the time series as a point cloud using tools from time series analysis, and then represent the data using topological features extracted from the shape of the resulting point cloud.
The features are extracted from persistence diagrams (a tool from TDA) which provide a summary of the topological features in the point cloud.
We use the described approach to obtain a classification for chatter in a deterministic turning model, then apply the trained algorithm to a stochastic turning model.
The method's accuracy in classifying the chatter/chatter-free cuts in the deterministic case is determined using the stability calculations of the spectral element approach \citep{Khasawneh2011}.
For the stochastic model, we qualitatively comment on the resulting classification in light of the expected behavior of stochastic systems since the tools for the stability analysis of time-periodic, stochastic delay differential equations are slim or non-existent.
This inductive learning is especially useful when it is desirable to train the signal on an abstract system model where the ground truth is known, and then hope to transfer the acquired knowledge to a more complicated system where it is more difficult to classify the resulting behavior.


\section{Mechanical Model}
\label{sec:Model}
\begin{figure}[htbp]
 \centering
 \includegraphics[width=0.33\textwidth]{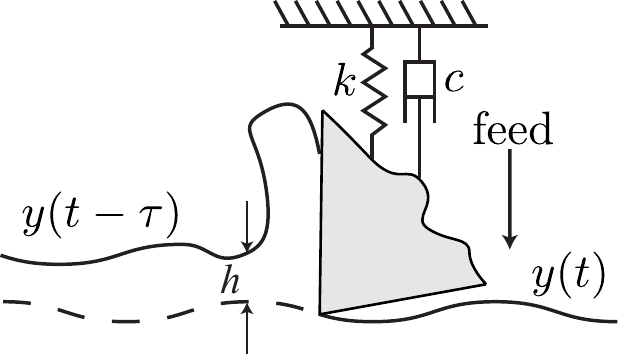}
\caption{The turning model described in Sec.~\ref{sec:Model}.}
\label{fig:sdofturning}
\end{figure}

Figure~\ref{fig:sdofturning} shows the turning process that will be investigated in this study.
The tool is modeled as a linear oscillator with a single degree of freedom in the $y$ direction while the workpiece is assumed to be rigid. The cutting force $F$ acts at the tool tip and the resulting equation of motion reads
\begin{equation}
 \ddot{y} + 2\zeta \omega_n \dot{y} + \omega_n^2 y = \frac{F}{m}=\frac{K w h^{\alpha}}{m}
 \label{eq:sdof1}
\end{equation}
where $\zeta$ is the damping ratio, $\omega_n$ is the natural angular frequency of the tool, and $m$ is the modal mass;  $K$ is the mechanistic cutting coefficient with units of N/m$^2$ which relates the cutting force to the area of the uncut chip $A=w\times h$, $w$ is the depth of cut shown in Fig.~\ref{fig:sdofturning}, $h$ is the uncut chip thickness, and the exponent is typically chosen as $\alpha=0.75$ (\cite{Tlusty2000,Stepan2001a}).

The tool will oscillate under the influence of the cutting force leaving behind an undulated surface on the workpiece. When the tool then advances into the workpiece during the following revolution at the nominal feedrate $h_0$ meter per spindle revolution (m/rev), it will encounter the wavy surface from the previous revolution.
This dynamic variation of the dynamic chip thickness is thus influenced by both the current and the previous tool oscillations represented by $y(t)$ and $y(t-\tau)$, respectively.
Here $\tau = 2\pi/\Omega$ is the duration of one spindle revolution corresponding to the spindle speed $\Omega$ rad/sec.
The coupling between the cutting forces and the tool oscillations can lead to large amplitude, self-regenerative vibrations referred to as chatter.
If the tool oscillations are severe, then the tool will leave the surface of the workpiece and experience damped oscillations, i.e., the right hand side of Eq.~\eqref{eq:sdof1} becomes zero.

To reduce the number of parameters in Eq.~\eqref{eq:sdof1}, we use the rescaling described in \cite{Insperger2008a}: let $y(t) = h_0 \tilde{y}(t)$ and let $\tilde{t} = \omega_n t$, and $\tilde{\tau} = \omega_n \tau$.
After dropping the tildes we obtain
\begin{align}
 \label{eq:sdof2}
\nonumber \ddot{y} + 2\zeta  \dot{y} +  y &= \frac{K w (2\pi R)^{\alpha-1}}{m \omega_n^2} \rho^{\alpha-1} (1+y(t-\tau)-y(t))^{\alpha} \\
                                               &= b \rho^{\alpha-1} (1+y(t-\tau)-y(t))^{\alpha}
\end{align}
where $R$ is the radius of the workpiece, $\rho = h_0/(2\pi R)$, $b$ is the dimensionless depth of cut, and $\rho < 0.01$.
Equation \eqref{eq:sdof2} is used in the simulation to generate time series for the deterministic system.

We can capture the effect of the variations in the workpiece material, shear angle, or temperature effects, by introducing the stochastic non-dimensional cutting coefficient $\hat{b}$.
If we allow the dimensionless cutting coefficient to become a stochastic variable
\begin{equation}
\hat{b} = \bar{b} + \delta \frac{dB}{dt}
\label{eq:Kstoch}
\end{equation}
where $\bar{b}$ is the average or nominal value of the stochastic non-dimensional cutting coefficient $\hat{b}$, $\delta$ is the diffusion coefficient, and  $B$ is standard Brownian motion.
The coefficient $\delta$ represents a measure of the stochasiticity in the system.
The resulting stochastic delay differential equation for the tool oscillations then reads
\begin{align}
 \label{eq:stoch}
 \nonumber d{\dot{Y}} =& \left( -2\zeta  \dot{Y} -  Y + \bar{b} \rho^{\alpha-1}  (1+Y(t-\tau)-Y(t))^{\alpha} \right) dt \\
  &+ \delta  \left( \rho^{\alpha-1}  (1+Y(t-\tau)-Y(t))^{\alpha} \right) dB
\end{align}
where Eq.~\eqref{eq:stoch} is interpreted in the It\^{o} sense \citep{Oksendal2007}.
We use Eq.~\eqref{eq:stoch} in this paper to generate the noisy time series.


\section{Topological Signal Processing}

In order to automate the classification of the time series arising from our model, we turn to the new field of topological signal processing, which combines methods from Topological Data Analysis (TDA) with tools from time series analysis to analyze streams of data.

\begin{figure}
 \centering
 \includegraphics[width = .2\textwidth]{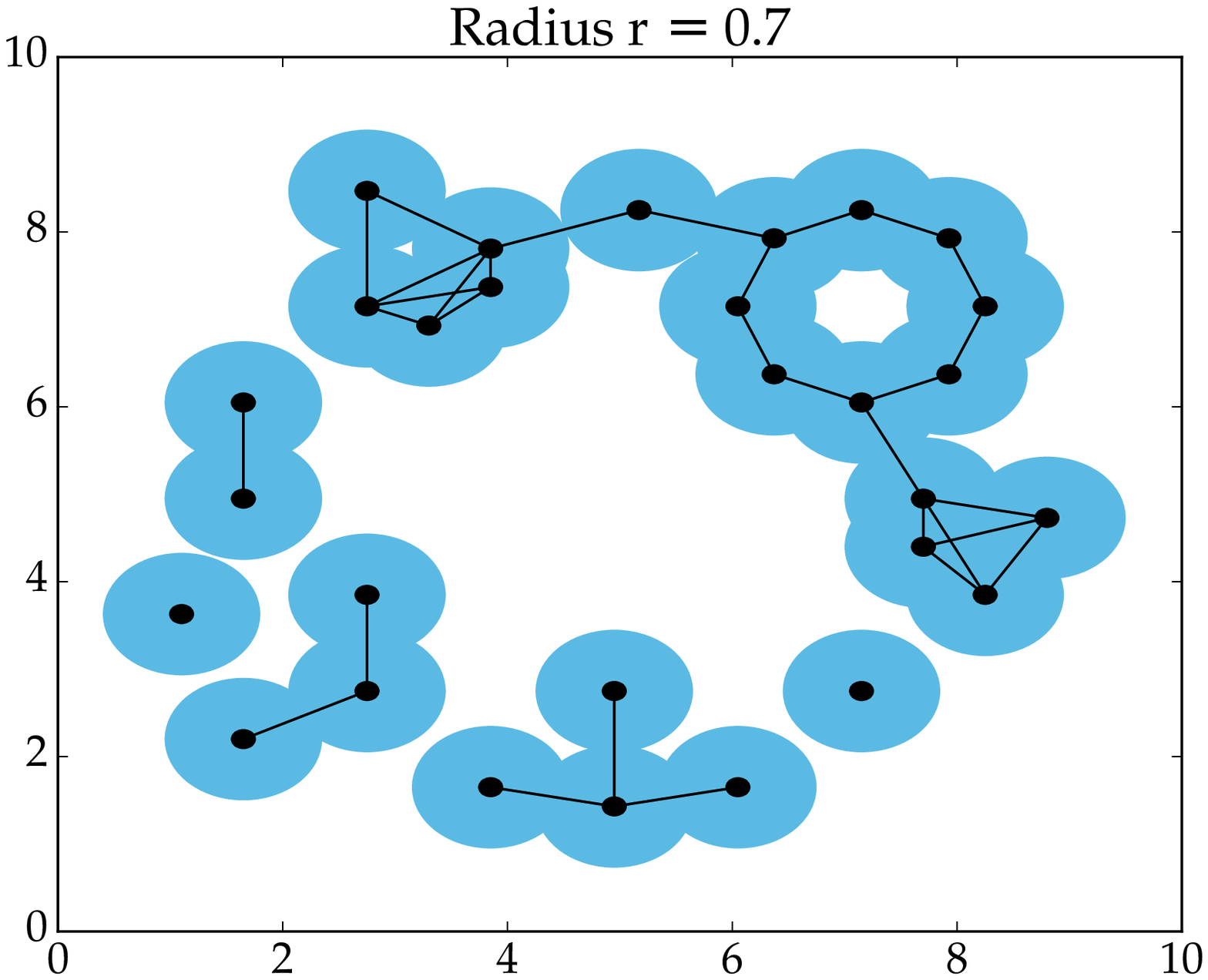}
 \includegraphics[width = .2\textwidth]{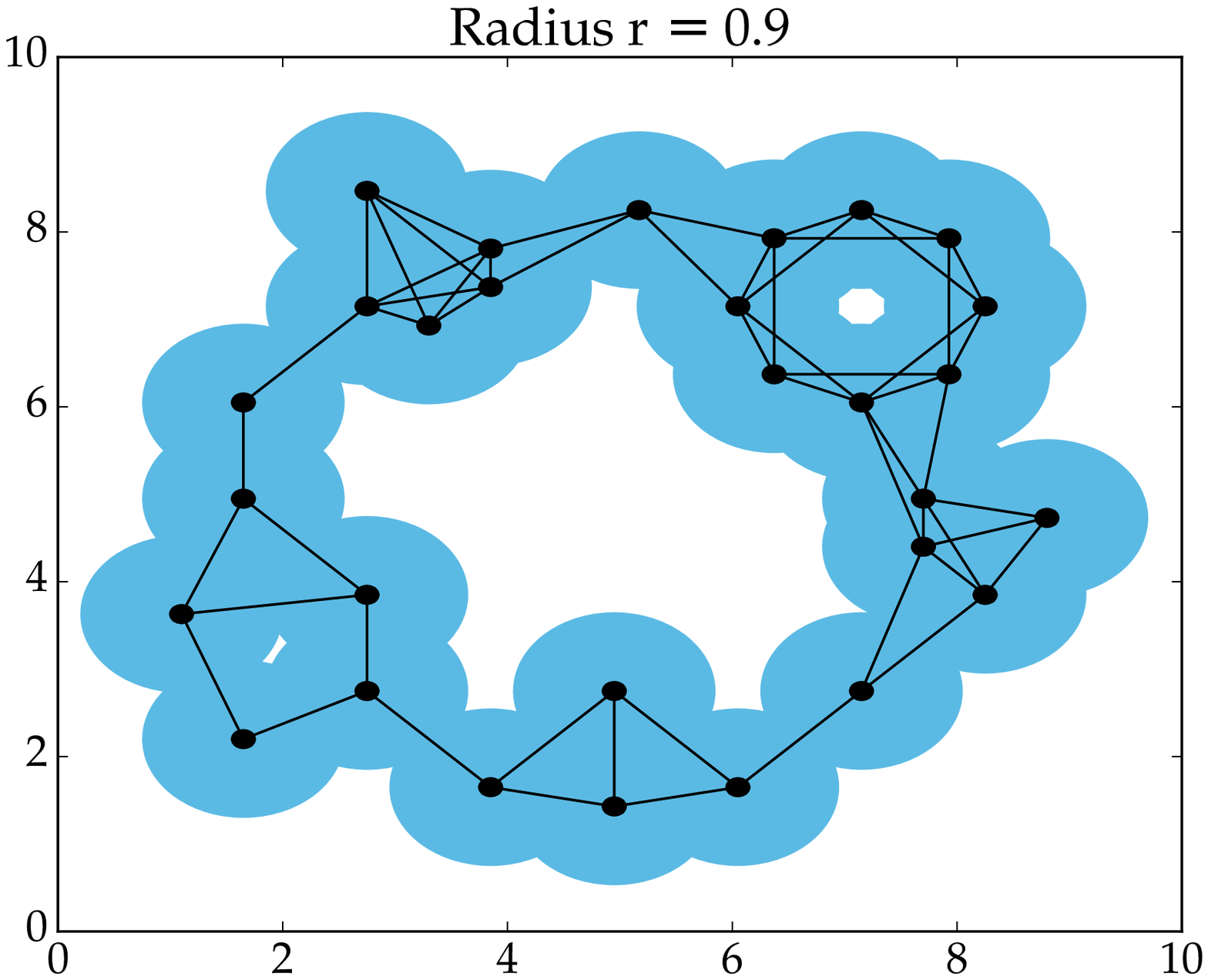}
 \includegraphics[width = .2\textwidth]{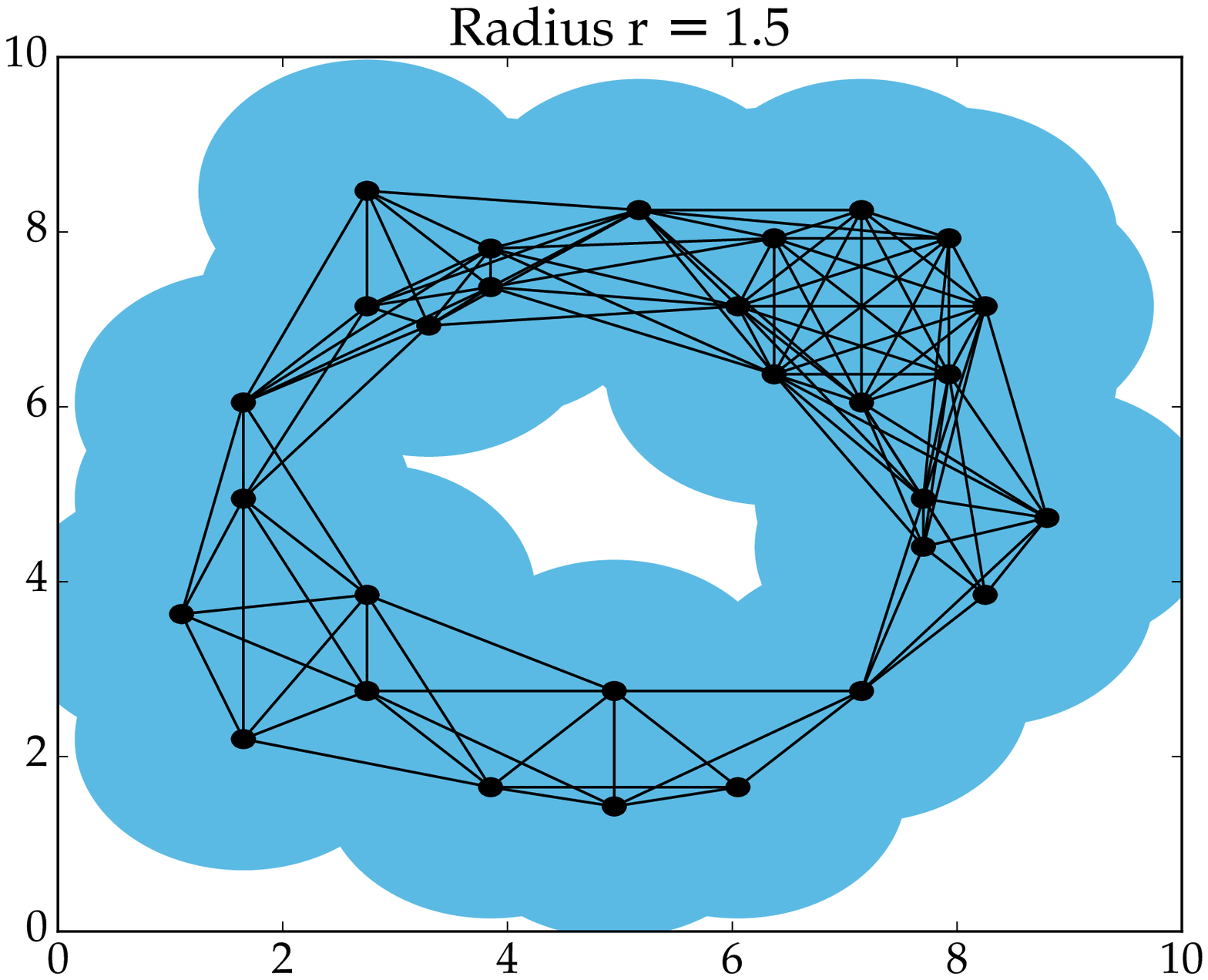}
 \includegraphics[width = .2\textwidth]{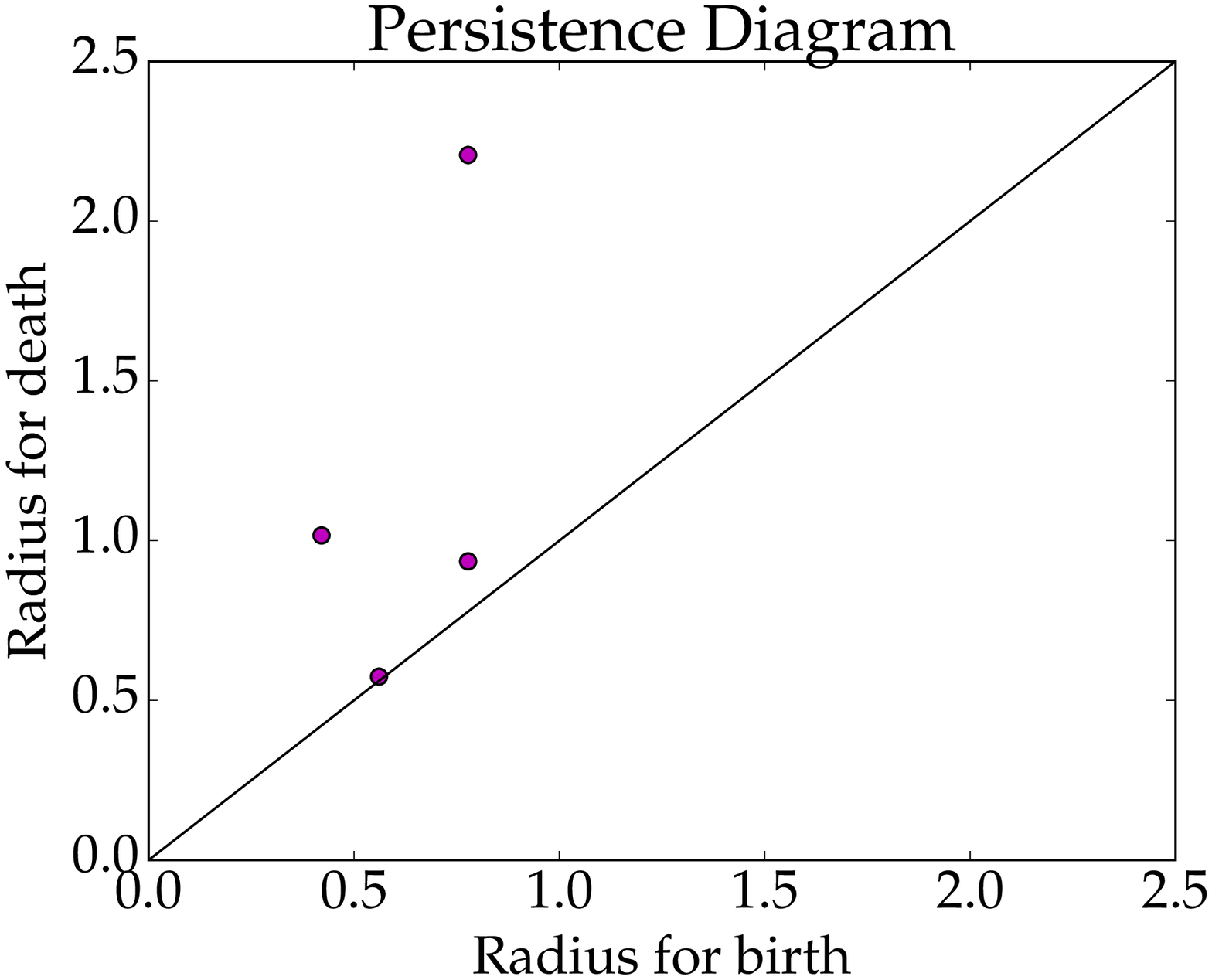}
\caption{A point cloud shown with collections of blue disks for several different radii (top left, top right and bottom left); the persistence diagram for this point cloud is given in the bottom right.  The circular structure of the point cloud is reflected in the single point far from the diagonal in the persistence diagram. }
\label{fig:Circle}
\end{figure}

\subsection{Persistent Homology}
We begin with an informal introduction to the subject of persistent homology, and direct the reader to \cite{Munch2017} or \cite{Edelsbrunner2010} for a more in depth discussion.
Suppose we are given a point cloud drawn from a manifold and want to understand something about the underlying structure.
To do this, we consider expanding a collection of discs of the same radius centered at each point.
We can then study the structure of the union of these discs for a changing radius.
In the example of Fig.~\ref{fig:Circle}, we start with a point cloud sampled from an annulus and see that at a very small radius (around $r=0.3$), the collection of blue disks consists of a set of disconnected components.
At a slightly larger radius ($r=0.7$), these discs start to intersect, possibly forming small circular structures which fill in at a still slightly larger radius.
What is very interesting is that at a relatively small radius (approximately $r=0.9$), we form a circular structure representing the full annulus which takes a much longer time to fill in than the small circular structures seen previously.
The goal of persistent homology is to quantify this intuition in a rigorous manner and use it to ``measure'' the size of the annulus.

A hole is called a ``class'' in the homology.
While homology gives information about a static topological space, persistent homology quantifies the changing homology of a changing topological space.
In particular, persistent homology measures how long each class persists in the changing space.
Having a class with a long life  quantifies our intuition that the point cloud appears to have circular structure.

To make this mathematical construction computable, we approximate the structure of the union of disks by a simplicial complex.
A simplicial complex is a topological space built up from discrete, combinatorial building blocks.
It can be thought of as a generalization of a graph since the lowest dimensional building blocks are vertices and edges; however, it can also contain higher dimensional simplices such as triangles, tetrahedra, and their higher dimensional analogues.
The edge set of this simplicial complex is drawn in Fig.~\ref{fig:Circle} as black lines between the points; we assume that higher dimensional simplices are included whenever all their lower dimensional faces appear.
By representing the changing space as a combinatorial structure, we have the ability to store its data in a computer as well as to compute its persistent homology and keep track of the appearance and disappearance of classes.
The persistence diagram (i.e.~bottom right of Fig.~\ref{fig:Circle})  plots the lifetime of each of these classes by placing a point for each class at $(b,d)$ if the class appeared (was born) at radius $b$ and was filled in (died) at radius $d$.
In the example of Fig.~\ref{fig:Circle}, small loops that appear and disappear quickly in the set of expanding discs, correspond to points close to the diagonal in the persistence diagram.
The large circular structure appears as a point far from the diagonal.
In this way, we have a visual representation of the difference between classes which live a long time  (and could be considered important), and the points representing short lived classes (and are often assumed to be noise).
For this reason, we usually draw the diagonal $ \{(x,x)\mid x\geq 0 \}$ into the persistence diagram along with the off-diagonal points to  remind the observer  that points always appear above the diagonal, and that proximity to the diagonal implies that an off-diagonal point is more likely to be noise.
While this example shows that our intuition is valid in the case of a 1-dimensional structure (the circle) embedded in visualizeable 2-dimensions, the persistence diagram can give information about a structure of any dimension embedded in any high dimensional space, while still remaining as a 2-dimensional output.

\begin{figure}
 \vspace{-.3in}
\begin{center}
\includegraphics[width = .5\textwidth]{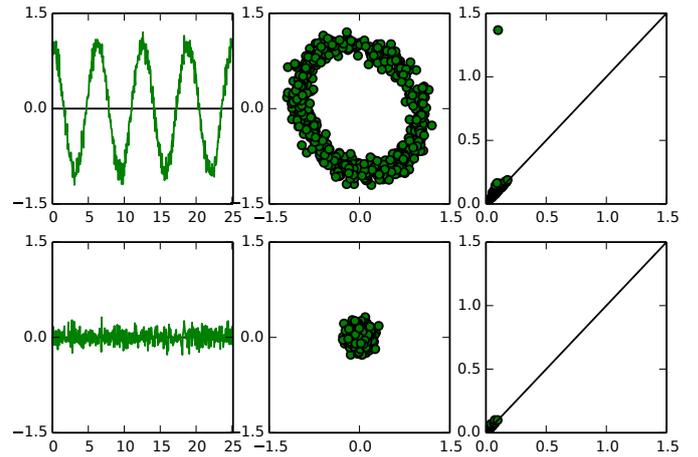}
\end{center}
\vspace{-0.2in}
 \caption{
 An example of the relationship between periodicity and maximum persistence.
 The columns correspond to the signal, the Takens embedding of the signal, and the persistence diagram of the Takens embedding, respectively.
 Note that the periodic signal has a prominent off-diagonal point in the persistence diagram unlike the signal that only contains noise.
 }
 \label{figs:signalCompare}
\end{figure}


\subsection{Delay Embedding and Persistence}
\label{sec:delayAndPersistence}

The delay embedding is a standard tool for time series analysis \citep{Kantz2004}.
Given a time series $X(t)$, which in practice is a set of samples $s_n = X(t_n)$,
fix a delay $\eta>0$ and choose a dimension $m \in \Z_{>0}$ in which to embed the data.
Then the delay embedding is a lift of the time series to the map to $\R^m$
\begin{equation*}
 \psi_\eta^m: t   \longmapsto  (X(t),X(t+\eta), \cdots, X(t+(m-1)\eta)).
\end{equation*}
Through an important theorem of Takens \citep{Takens1981}, we are justified in using the term ``embedding'' since, under the correct parameter choices, this mapping preserves the structure of the underlying manifold and the dynamics of the system.
%

The discrete representation of the time series $X(t_n)$ leads to a point cloud $\{\psi_\eta^m(t_n)\}$ called the Takens embedding.
In the case that a time series contains periodicity, the Takens embedding is a point cloud with a circular structure.
Of course, we can compute the 1-dimensional persistence of the generated point cloud in order to measure the size of this structure.
Circularity in the point cloud is quantified in the persistence diagram as a point far from the diagonal as shown in the first row of Fig.~\ref{figs:signalCompare}.
If the time series is just noise, there is no prominent off-diagonal point which can be seen in the second row of Fig.~\ref{figs:signalCompare}.
For a theoretical treatment of  persistence and delay embeddings see \cite{Perea2015} and \cite{perea2016persistent}.

\subsection{Featurization of Persistence Diagrams}
\label{sec:Featurization}

Persistence diagrams encode a great deal of information about the structure of a point cloud, making it very useful for data analysis.
However, because the space of persistence diagrams does not have an inner product structure, direct application of most machine learning algorithms is not possible.
Thus, we turn to the featurization of persistence diagrams for help; that is, the construction of a map which returns a point $ \phi(D) \in \R^N$ for a given persistence diagram $D$ stored as a list of its off diagonal points.

For this work, we choose the features of \cite{Adcock201} combined with maximum persistence.
The basic idea of this method of featurization is to work with polynomials defined on the collection of off-diagonal points, but which respect the structure inherent in the persistence diagram.
First, these polynomials must be able to accept persistence diagrams with different numbers of off-diagonal points since even beginning with input data containing the same number of points, the persistence diagrams can be vastly different in size.
Secondly, while a persistence diagram can be stored as its list of off-diagonal points, $D = \{(x_1,y_1),(x_2,y_2),\cdots,(x_n,y_n)\}$, the order of presentation does not change the diagram, so any given function must present the same output no matter the input order.
To this end, we choose the following collection of functions,
\begin{equation*}
\begin{array}{rl}
f_1(D) & =  \sum{x_i(y_i-x_i)}, \\
f_2(D) & = \sum{(\overline y - y_i)(y_i - x_i)},\\
f_3(D) & = \sum{x_i^2 (y_i - x_i)^4},  \\
f_4(D) & = \sum{(\overline y - y_i)^2 (y_i-x_i)^4},\\
f_5(D) & = \max\{ (y_i - x_i) \}
\end{array}
\end{equation*}
where $\bar y$ is the maximum death time, and
the summations and maximum are each taken over all points  in $D$.


\section{Numerical Simulation}
\label{sec:NumSimulation}
Two sets of simulations were used: MATLAB's DDE23 command for the deterministic case in Eq.~\eqref{eq:sdof1}, and an Euler-Maruyama simulation for the stochastic system of Eq.~\eqref{eq:stoch}.
For the stochastic case, multiple realizations of the solution path were generated for use in the analysis.

\begin{figure}[htbp]
	\centering
	\includegraphics[width=0.45\textwidth]{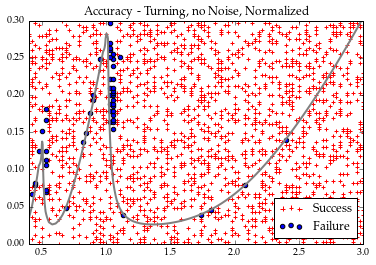}
	\caption{Accuracy plot in turning with no noise. }
	\label{fig:accuracy_plot_turning}
\end{figure}

The system parameters that were used in the simulations are $\zeta=0.03$, $\rho=0.01$, and $\alpha=0.75$, and the simulation was performed over a $100 \times 100$ grid in the $(\Omega/\omega_n,b)$ space.
To provide a basis of comparison for the analysis of both the deterministic an the stochastic models, we also include the results of the linearized stability analysis of the noise-free model~Eq.~\eqref{eq:sdof1}.
There are several methods available in the literature to do this, e.g.~zero-order approximation \citep{Altintas1995}, semi-discretization \citep{Insperger2004}, Chebyshev collocation \citep{Butcher2011}, extended Tlusty's law method \cite{Otto2014}, or spectral element method \citep{Khasawneh2011}.
In this study, we use the latter approach with one temporal element, a polynomial of degree 12, and a $100\times100$ grid in the $(\Omega/\omega_n,b)$ plane (see \cite{Khasawneh2011} more details).
The resulting stability diagram is the  gray line in Fig.~\ref{fig:accuracy_plot_turning}.
The region below the line represents the chatter-free regime while the region above marks the chatter regime.

The stochastic case, Eq.~\eqref{eq:stoch} was simulated using the Euler-Maryuama method \citep{Buckwar2000}, while the Brownian path was created using the approach described in \cite{Higham2001}.
For more details on the method and parameters, please refer to \cite{Khasawneh2015}.

\section{Methods, Results, and Discussion}
\begin{figure*}[h!]
	\centering
	\includegraphics[height = 4.4cm, trim={0 0 0 0},clip]{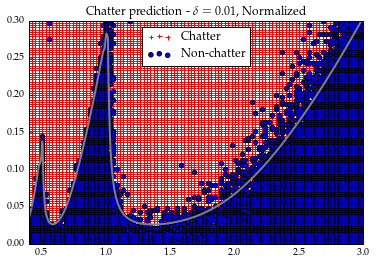}
	\includegraphics[height = 4.4cm, trim={1cm 0 0 0},clip]{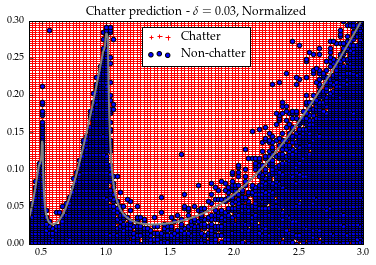}
	\includegraphics[height = 4.4cm, trim={1cm 0 0 0},clip]{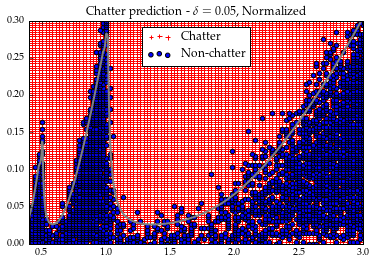}
	\caption{The results of classification on the data from the stochastic model using the classifier trained on the deterministic model. The stability boundary is included in gray for comparison, but is not included in the calculations.}
	\label{fig:NoiseCases}
\end{figure*}

Following the computations of \cite{Khasawneh2015}, we used persistent homology  to analyze the time series generated by the numerical simulation.
For every experiment, each time series
was given as a series of samples $s_n = X_{\Omega/\omega_n,b}(t_n)$ for $t_n \in [0,T]$, where $T$ depends on the value of $\Omega/\omega_n$ according to $ \frac{2^5 \times 2 \pi}{\Omega/\omega_n}$.
The second half of the time series was retained to avoid transients, and was sub-sampled to speed up persistence computations.
This resulted in $264$ remaining data points evenly distributed in $[T/2,T]$ for all the time series, both with and without noise.

The time lag $\eta$ was determined for each time series using the first zero of the autocorrelation function.
This was used to construct the Takens embdding as a point cloud in $\R^3$.
We computed the 0- and 1-dimensional persistence diagrams for each Takens embedding using the M12 package \citep{Harer2014}.
For each of the diagrams, we computed the five functions described in Sec.~\ref{sec:Featurization}.
Because the 0-dimensional persistence diagram for the Rips complex of a point cloud has all birth times equal to 0 in the 0-dimensional persistence diagram, $f_1$ and $f_3$ were entirely 0, so these features were discarded.
This resulted in a total of 8 input features computed for each parameter value which were normalized prior to the analysis.
Machine learning was done using the \texttt{sklearn} python package.
Specifically, a logistic regression classifier\footnote{\texttt{sklearn.linear\_model.LogisticRegression} with default inputs} was trained on 80\% of the time series and tested using the remaining 20\%.
The labels for classification were determined using the stability diagram of the deterministic model (Eq.~\ref{eq:sdof1}).

Using the featurization of the persistent homology for machine learning resulted in a 97\% classification rate.
The confusion matrix for the classification test on the noise-free data set is
\begin{equation*}
    \begin{blockarray}{rccc}
     & & \multicolumn{2}{c}{True}\\
      &   & \text{Chatter} & \text{Not-Chatter}  \\
      \begin{block}{rr(cc)}
       \multirow{2}{*}{\rotatebox[origin=c]{90}{Pred.}} & \text{Chatter} & 1250 & 46 \\
       & \text{Not-Chatter} & 6 & 698\\
      \end{block}
    \end{blockarray}
\end{equation*}
and the locations of the misclassified  time series can be seen in the blue dots of Fig.~\ref{fig:accuracy_plot_turning}.
Notice that not only is the classification rate incredibly high, the locations of the misclassified time series are restricted to the boundary of the stability matrix.
As the stability diagram was calculated numerically on a discrete mesh, we suspect that issues with meshing and possibly with machine precision have led to a slight lowering of the two sharp peaks in the boundary.
If this is true, the machine learning model is actually recreating this peak but interpreting it as a misclassification since our labels are constructed using the calculated boundary.

We then used the trained logistic model on the data sets arising from the stochastic model (Eq.~\eqref{eq:stoch}).
That is, using the same procedure for generating the Takens embedding and computing the normalized 8-dimensional feature vector, we classified the resulting output.
The results of the classification can be seen in Fig.~\ref{fig:NoiseCases}.

Notice in particular that the labeling implies that small amounts of noise (i.e., $\delta = 0.01$) actually increases the stability of the system since more parameters are considered to have a non-chatter output.
This can be attributed to stochastic resonance (\cite{Kuske2010}).
Meanwhile, we can see that as the noise increases, (i.e., $\delta = 0.03$ and $0.05$) the chatter labeled parameters increasingly infiltrate the non-chatter region, thus leading to more instability.
Nevertheless, the approach we describe can provide chatter classification for signals produced by complicated and noisy manufacturing systems especially when other methods cannot be applied.



\end{document}